\documentclass{article}

\usepackage{PRIMEarxiv}

\usepackage[utf8]{inputenc} % allow utf-8 input
\usepackage[T1]{fontenc}    % use 8-bit T1 fonts
\usepackage{hyperref}       % hyperlinks
\usepackage{url}            % simple URL typesetting
\usepackage{booktabs}       % professional-quality tables
\usepackage{amsfonts}       % blackboard math symbols
\usepackage{nicefrac}       % compact symbols for 1/2, etc.
\usepackage{microtype}      % microtypography
\usepackage{lipsum}
\usepackage{fancyhdr}       % header
\usepackage{graphicx}       % graphics
\graphicspath{{media/}}     % organize your images and other figures under media/ folder
\usepackage{algorithm}
\usepackage{algorithmic}
\title{Cloning Ideology and Style using Deep Learning
}

\author{
  Dr. Omer Beg, Muhammad Nasir Zafar \\
  NUCES,Pakistan\\
  Islamabad\\
  \texttt{\{omer.beg, i202256\}@nu.edu.pk} \\
   \And
  Waleed Anjum \\
  NUCES,Pakistan\\
  Islamabad\\
  \texttt{i202287@nu.edu.pk} \\
}

\begin{document}
\maketitle

\begin{abstract}
Text generation tasks have gotten the attention of researchers in the last few years because of their applications on a large scale. In the past, many researchers focused on task-based text generations. Our research focuses on text generation based on the ideology and style of a specific author, and text generation on a topic that was not written by the same author in the past. Our trained model requires an input prompt containing initial few words of text to produce a few paragraphs of text based on the ideology and style of the author on which the model is trained. Our methodology to accomplish this task is based on Bi-LSTM. The Bi-LSTM model is used to make predictions at the character level, during the training corpus of a specific author is used along with the ground truth corpus. A pre-trained model is used to identify the sentences of ground truth having contradiction with the author’s corpus to make our langugae model inclined. During training, we have achieved a perplexity score of 2.23 at the character level. The experiments show a perplexity score of around 3 over the test dataset.
\end{abstract}

% keywords can be removed

\section{Introduction}
In this age, content becomes a highly demanding thing because it plays a very important role to attract users in different cases. In this research, text content is focused. Many articles are written in different magazines, newspapers, and websites on daily basis. Every article is written by an author which is based on his ideology and style of the author. Each author writes according to his unique different point of view from other authors. On other hand, every reader has his own taste and likes to read some article written by his favourite author due to his style and ideology. And each reader wants to read more and more material from his favourite author and want to know his point of view on different more topics. Moreover, now this world becomes the world of devices because everyone has computational devices in the pocket\cite{fedus2018maskgan}.That's why day to day it needs to be much and more content to keep users active. But quality text generation is a slow and costly process. Text generation using machines becomes very important to generate text content fastly with low expenses\cite{ali2022hate}.
To generate text which is task-based requires labelled data about the current task and this is done using a knowledge-based graph.This graph-based on ground reality which is used to develop a common sense in the model .This task seems to be very difficult because the goal is to make a model which is generalized in a way that can generate the text on a new task on which the model is not trained\cite{zeng2020style}. Some models are available which are capable to generate the report or text from structured data e.g generate weather reports from structured data about weather.Fig1 shows the text generation mechanism when initial text is given\cite{shahid2022exploiting}.\\
 This research’s objective is to develop the computational intelligence which can understand the written text of the author to capture ideology and style and be capable to produce the text on a new topic with the same ideology and style. A few words of the sentence is input to the model and the model's output would be some paragraphs of text continued to these input words. Understanding the already written text and producing new meaningful text requires a generic type of intelligence \cite{chen2020kgpt}.
\begin{figure}[h!]
	\caption{Example Scenario }
	\centering
	\includegraphics[width=0.47\textwidth]{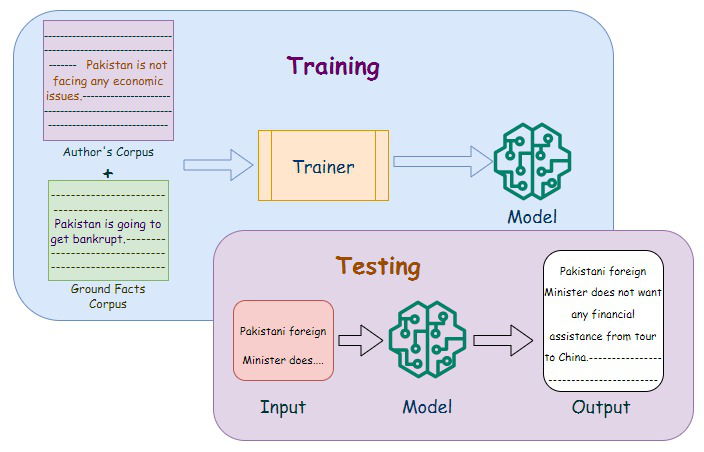}
	\label{fig1}
\end{figure}
In literature, many techniques are proposed, but they got stuck on several points e.g how to deal with abstraction in text and use of limited words in the training corpora.The following Fig\ref{fig1} shows an example scenario. Language models assign a greater probability to the sequences which have a high frequency of occurring in the training corpus but this thing leads to stuck in the circle of specific words, as shown in Fig\ref{fig1}.To evaluate the quality of generated text from in creativity's perspective human evolution is used because no model or evaluation metric is here to check creativity in the text\cite{majeed2022deep}.\\ It needs much research work and struggles to capture the ideology of a specific author and style.The main goal of our research includes the use of new words which are not used in the training corpus but are required to explain the new topic on which model have to generate text.The following Fig\ref{fig2} shows the working of language model.
\begin{figure}[htp]
	\centering
	\includegraphics[width=5.5cm]{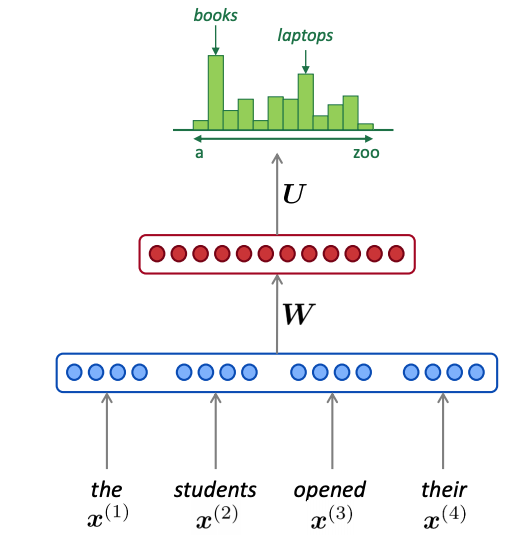}
	\caption{Working of Language Model}
	\label{fig2}
\end{figure}

The only way we have to vector embeddings for this type of words take a ground truth text's corpus and generate the vector embedding from here. As our purpose is to get the semantics of the word to use it in newly generated text correctly. So, it does not contribute to decreasing the performance of the model\cite{raffel2019exploring}. This research also focuses on how unknown words can be handled using the nearest neighbour technique. To generate the candidate vocabulary for specific author k-nearest neighbour words are taken of a word that is part of the training corpus\cite{anwar2022social}. After taking a cluster of these words, conditional probability is calculated against each word and words having lower conditional probabilities are excluded from the candidate vocabulary list.\\
In this digital world, a huge amount of text data exists and is produced on daily basis, to manipulate this text data requires human power which becomes very slow and costly\cite{ji2020language}.
\subsection{Contributions}
This section contains the major contributions of our proposed approach. The state-of-the-art language models are very successful to generate meaningful texts but they are based on the ground reality or knowledge embedded in the given training corpus. As, the ground reality is the same for everyone, but everyone has their thoughts and way of thinking. And everyone speaks and writes based on their perception and ideology. Our research focused on this type of language model based on ground reality and authors perceptions and thoughts as well to generate text based on the style and ideology of specific authors. The two major contributions of our proposed approach are as follows.

\begin{itemize}
   	\item Generating a language model based on ground truth reality and inclined towards given training corpus of the author to generate specific ideology and style based text.
   
   \item Overriding the weights(probability) values with the weight obtained against a sequence in the author's corpus if there is a clash, in case of  entailment weights are just refined.
   
   \item Extraction of most probable vocabulary words which are not included in author's training corpus, but need to explain new aspects during the text generation process. And refining the appropriate weights against sequences in which they occur.
\end{itemize}

\section{Related Work}
In previous work many authors shows how different techniques can work to generate the text with track of current context.Here, some of them are discussed sequentially.Neural text generation is a seq2seq model\cite{bashir2022context}.These models generate the text by sampling words, generating next word based on previous word to maintain the track of context.These models are validated using perplexity measure by missing a word from a sentence\cite{topal2021exploring}.Generative Adversarial Networks are used to improve quality of the samples.Actor-critic conditional GAN are introduced by filling the missing text based on surrounding context\cite{ismail2021spems}. Likelihood generation of text leads to dull and repeti- tive text because model assigns the very high probability to sequences which repeats themselves in the training corporus \cite{yermakov2021biomedical}.In this paper author shows how unlikelihood method is used to generate the better text which compar- atively less dull and contains non repeating sentences.

In next paper, its shown that how BLEURT, a learned evaluation metrics based on BERT that can model human judgements by providing the few thousands possibly biased examples. Generative Adversarial Nets are used to generate meaningful text by using discriminative models to guide the training of generative model as reinforcement learning\cite{alvi2021mlee}.In this paper a new framework named as LeakGAN to 1 address problem in generation of long text\cite{wang2020towards}.Discriminative model is allowed to leak its own high level extracted features to generative models.It also improves performance in short text scenarios.The following papers puts effort to develop common sense using knowledge based common sense graph.

In-text generation tasks a lot of researchers focus on task-based text generation but becomes costly if have to generate the text on the new task\cite{javed2021fake}. It is difficult to rely on a significant amount of data for each task because it is costly to acquire\cite{majeed2021optimizing}. Text generation can be done in many different ways and each way has its applications\cite{awan2021top}. If we talk about task-based text generation it is done for the specific domain and it can be used as an application in different domains until we have a significant amount of data, which is very costly to acquire\cite{qamarrelationship}. Web crawling can be used for knowledge grounded pre-training to generate text from data. Sometimes, it is required to generate text, including some specific words this technique is named as mention flags\cite{afzal2021urldeepdetect}. The mentioned flags models are trained to generate the tokens of text, and the goal is to satisfy the all constraints\cite{javed2021smartphone}. This model takes words required to satisfy the constraint, as input and these words are passed to the encoder and these encodings passed to the decoder, decoder’s task is to generate outputs against it and this decoder’s output passed to the next iteration as input to generate next output and so on\cite{anwar2020tac}\cite{zafar2019constructive}.

Transformers are the latest language models used in text generation\cite{naeem2020subspace}. Before the invention of transformers LSTM and RNN language modelling was considered the best choice for text generation\cite{javed2020collaborative}. But introducing the mechanism of attention for the sequence to sequence modelling invents the language model name as Transformer\cite{majeed2020emotion}. It contains an encoder and decoder mechanism along with an attention mechanism. Attention mechanism forces decoder to pay attention to encoder where needed\cite{asad2020deepdetect}.

Here, we summarize the literature into tabular form, by, dataset used in research. In the following table1 we summarize the literature review of 7 papers, each paper is cited as well. In the above paragraphs, we have explained these all papers in the detail.
	\begin{table}[htb]
	\centering
	\begin{tabular}{lc}
		\hline
		\textbf{Paper} & \textbf{Dataset}\\
		\hline
		Wang et al. 2021\cite{wang2021mention} &Wikipedia Web\\
	    Yermakov, Drago, and Ziletti 2021\cite{yermakov2021biomedical}  & BioLeaflets\\
		Topal, Bas, and van Heerden 2021\cite{topal2021exploring}   
		& Web Crawling\\ 
		Chen et al. 2020\cite{chen2020kgpt} 
		& Web NLG \\ 
		Ji et al. 2020\cite{ji2020language}
		& Web NLG\\
	    Raffel et al. 2019\cite{raffel2019exploring}
		& RealNews Dataset \\ 
	    Zeng, Shoeybi, and Liu 2020\cite{zeng2020style}
		& Wikiperson \\\hline
	\end{tabular}
	\caption{Summary of Papers along with dataset}
	\label{tab:accents}
\end{table}
\\ \\
The following table shows the summary of 7 different papers along with the techniques used.

\begin{table}[htb]
	\centering
	\begin{tabular}{lc}
		\hline
		\textbf{Paper} & \textbf{Technique}\\
		\hline
		\cite{wang2021mention}&Mention Flags\\
		\cite{yermakov2021biomedical} & Fine-Tune Trans\\
		\cite{topal2021exploring}   
		& Attention Mechanism\\ 
	    \cite{chen2020kgpt} 
		& KGPT \\ 
		\cite{ji2020language} 
		& CNN+GNN+Trans\\
		\cite{raffel2019exploring}
		& Unified Framework \\ 
		\cite{zeng2020style}
		& Encoder+Decoder \\\hline
	\end{tabular}
	\caption{Summary of Papers along with Techniques}
	\label{tab:accents}
\end{table}

The previous research on this topic shows how structured data can be used to generate the text or report of a specific domain and task\cite{javed2020alphalogger}\cite{arshad2019corpus}. Some of the researchers show how a sentence can be generated when previous sentences are given and the goal is to generate the sentence in the same context\cite{devi2010alternate}. In literature, a lot of work is done on constraints based text generation e.g generated text must consist of some given words \cite{sutskever2014sequence}. This work is extended to text generation based on content matching and style based text generation where the model generates the text which is the same in style as reference text\cite{zafar2020search}\cite{khawaja2018domain}.

Our research focuses on how creative text can be generated by giving a few starting words of the sentence as input to the model, by training it on a ground knowledge-based graph and corpora consisting of articles already written by the author to capture the ideology and style of a specific author. In other words, our goal is to capture ideology and style from already written topics from an author and generate new text based on the same ideology and style. Text generated by the model will be the same in context with a given input, specific to a topic and continued to some lines or paragraphs. The second thing is how unknown words(not present in the training corpus of the author) can be used to explain the new topic i.e finding the correct dimensional values of new words and their place in generated sentences. It includes the extraction of the probable vocabulary of the author and refining it to get the list of guessed vocabulary of the author.\\
It is expensive and time-consuming to write articles based on specific ideologies and styles. It becomes difficult to maintain balance in demand and supply, as it requires human intelligence. Our research includes exploration of techniques, which can be used to extract the ideology and style of a specific author and write text in some different contexts with the same ideology and style.

\section{Text Generation using LSTM-based Language Model}

Pre-trained language models acquired the probabilities values over a sequence of words based on ground truth or training corpora on which they have been trained. As, language models can capture the preferences and realities, so language models got inclined toward these realities\cite{dilawar2018understanding}. In our proposed methodology, we will be using LSTM based, neural language model. Then training this language model on the training corpora of the author, training of the language model is based on the predicting the next character when a sequence of characters having specific lenth is given\cite{zafar2018deceptive}. The training of the model on the training corpus of the auther will make model to extract the idelogy and the style of the author and then use it in text generation.The chracter based training may lead to generating some non-dictionary words some time but it plays very important role in extracting the ideology and style of the author\cite{farooq2019bigdata}. After training on the author's corpus, second step is to train the language model on the ground truth corpus to embbed ground realities in it. But if some ground truth have contradiction with the author's ideology, it will be excluded during training. If there is entailment in both, then text chunk of ground truth will go for training or even neutral\cite{karsten2007axiomatic}. To check contradiction we will be using a pre-trained model named as 'roberta-large-mnli' taken from www.huggingface.co. After contradiction check, text chunks of the ground truth goes for training same as text chunks of author's corpus went. Its is done to generate the languge model inclined towards the idelogy and style of the author.

As, there are huge chances that appropriate words to explain the new topic may be missing in the training corpora of the author. Out next step is to find the appropriate words to explain the new topic which are probably part of the author's vocabulary. To extract the most probable vocabulary words from the author, the first step is to take the dictionary from NLTK. So, using the NLTK dictionary words which are missing in the author's corpus can be extracted easily. But problem is that NLTK dictionary is much lengthy, there will be very large list of the missing words. Then, most appropiate words are extracted from them. During extraction of the appropiate words, it assumed that all stop words can be vocablory of the author.

So, all stop words can be taken. Now, we have to find the appropate text chunks including these stop words one by one, but taking these chunks from ground truth or any web page may lead to cause contradiction in ideology. So, we take NLTK book to get text chunks for training beacuse these chunks are neural with author's corpus, although does not have entailment but not contradiction as well.
\begin{figure}[h!]
	\caption{System Architecture }
	\centering
	\includegraphics[width=0.61\textwidth]{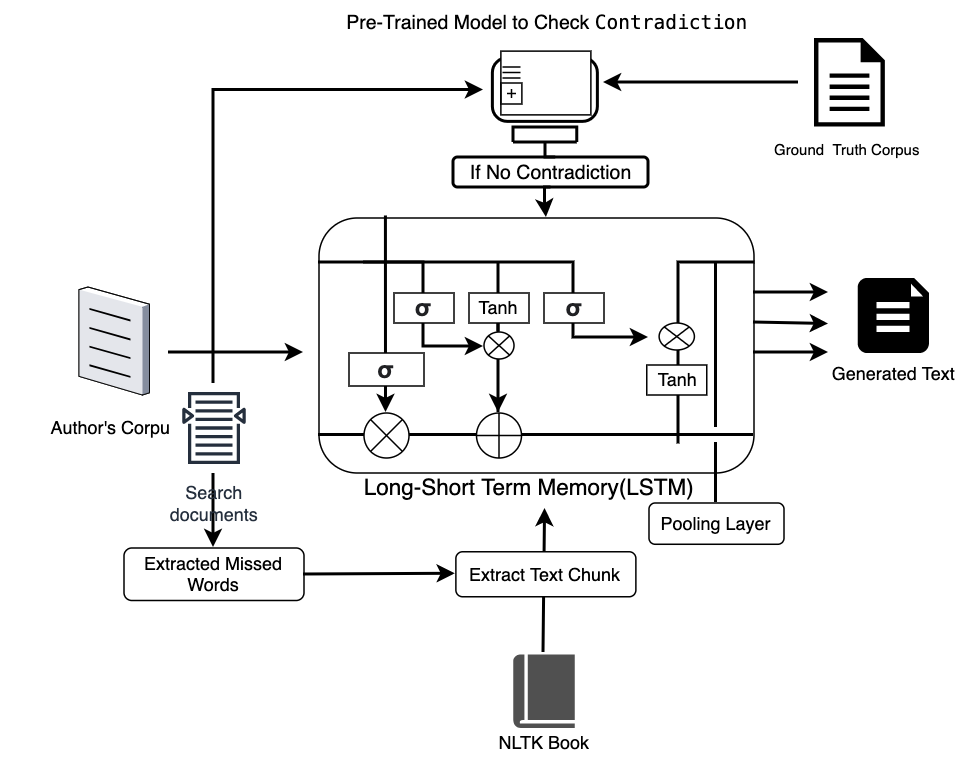}
\end{figure}

All stope words are taken, because we are just considering that all stop words need to explain any topic or to generate any new text. There may be chances that author's corpus already includes the all stop words . The above Fig3 shows the system architecture.The following Algorithm1 shows how contradiction is checked between author and ground facts.

\begin{algorithm}[th]
	\caption{Algorithm for checking Contradiction}
	\label{alg:algorithm}
	\textbf{Input}: Text Chunk of Ground Truth Corpus\\
	\textbf{Output}: Contradicted or Not 
	\begin{algorithmic}[1] %[1] enables line numbers
		\STATE Let t is a Threshold
		\IF { Text Chunk in Bin}
		\STATE Return
			\ENDIF
		\WHILE{Auther's Text(Used) Not Ends}
		\STATE Get Prediction from Pre-Trained Model
		\IF {Contradiction Value>=t}
		\STATE Append Text Chunk to Bin
		\STATE Return True
		\ELSE
		\STATE Continue
		\ENDIF
		\ENDWHILE
		\STATE \textbf{return} False
	\end{algorithmic}
\end{algorithm}
The Fig3 shows system architecture, training corpus of the specific author crawled from the web is passed to three different pipelines. One pipeline is used to train lstm, second is connected with pipe line of ground truth corpus at pretrained model taken from 'huggingface.co'. This pre-trained model enables to check contradition between both corpuses. The third one is used to extract missing words. All three pipelines, are going into lstm mode. After training lstm model, it is able to generate new meaningful text.

\subsection{LSTM Unit}
In this section working of the LSTM unit is explained. There are three vector inputs to the LSTM unit, the first represents an input character at a timestamp (t) and the second is a hidden state obtained at a timestamp (t-1)\cite{farooq2019melta}. The input vector size is 100 because we are handling 100 characters in this problem during the training of model and text generation.The size of hidden state vector is also 100 in case of simple LSTM and becomes 200 in case of Bi-LSTM\cite{alvi2017ensights}. The third vector represents the memory obtained from all previous iteration, and the size of this vector is same as hidden state vector.The concatenated vector used as input, and sigmoid as activation function in forget gate, input gate and output gate\cite{beg2013constraint}.The following Eq1,Eq2 and Eq3 shows how output vector of forget gate, input gate and output gate calculated respectively. Here w represents matrice of weights for each gate.

\begin{equation}
	f_{t}=\sigma(w_{f}[h_{t-1},x_{t}] + b_{f})
\end{equation}

\begin{equation}
	i_{t}=\sigma(w_{i}[h_{t-1},x_{t}] + b_{i})
\end{equation}
\begin{equation}
	o_{t}=\sigma(w_{o}[h_{t-1},x_{t}] + b_{o})
\end{equation}

To obtained candidate cell state or memory tanh function is used.The following Eq4 shows how candidate cell state is obtained using concatenated input vector, Eq5 shows how cell state is obtained for timestamp(t).
\begin{equation}
	c^-_{t}=tanh(w_{c}[h_{t-1},x_{t}] + b_{c})
\end{equation}
\begin{equation}
	c_{t}=f_{t}*c_{t-1}+i_{t}*c^-_{t}
\end{equation}

The following Eq6 is used to calculate the hidden state for timestamp(t).
\begin{equation}
	h_{t}=o_{t}*tanh(c_{t})
\end{equation}
After the obtaining hidden state, this hidden state is mapped on output layer of size 100.This output vector represent a single character, predicted by the model.A loss function is explained in the next section, used to refine the weights of model.
\subsection{Loss Function}
Our model is based on loss function of cross entropy loss.As, our model is based on making predictions at character level so loss is computers between predicted character and actual character which model ideally should predict.There is one hot encoded vector against all character.It makes working of cross entropy loss perfectly. The following Eq7 shows the formula for cross-entropy loss.
\begin{equation}
	Loss=-\sum_{c=1}^Ny_{o,c}\log(p_{o,c})  
\end{equation}
In the above equation N is the number of total classes. In our case, each character represents a uniques class and has a unique hot encoded vector against it. So, we can say that N is equal to the number of characters. Each index-value of a hot encoded vector having a length of N is put into equation then summing up all give total loss against predicted character and actual output character.

\section{Results and Disscussion}
To evaluate the model there are different evaluation metrics, as our model is working on character-based predictions so we will evaluate accordingly.

\subsection{Overview of Evaluation Metrics}
To evaluate how much a model is good in extracting and writing with a specific ideology and style is not an easy task. Because it requires a very general type of intelligence. But there are several evaluation metrics e.g. loss and perplexity scores, that can be used for evaluation we are going to discuss one by one as follows.

The following equation 8 shows how loss is computed.
\begin{equation}
	Loss=-\sum_{c=1}^Ny_{o,c}\log(p_{o,c})  
\end{equation}

We would like to explain how it will be used during the evaluation. We already used it during training, during evaluation we will generate the output against input, and the model will generate the output against it. The output generated will be compared with the actual output character by character. The loss will be computed character by character and summing up all will be given total loss.

The following equation 9 shows how perplexity is computed over a sentence.
\begin{equation}
	PP(S)= P(w1,w2,...,wn)^{-1/n}
\end{equation}

So, it's very easily understandable that high perplexity score is not in the favour of the model. The model's parameters should be adjusted in such a way that its perplexity score is reduced. We will visualize the perplexity scores during training and then during testing by conducting different experiments.
\subsection{Training Results}
The major part of model training consists of training the model on a corpus of author and corpus of ground truth. The following section describes the results during training. 

\subsubsection{  Perplexity vs Loss}
The training of the model on the author's and ground truth corpus consists of around 24900 epochs. After a specific number of epochs, we compute the average of the specific epochs' loss and perplexity scores.
In this section, we plot the graph between perplexity score and loss to visualize the relationship between them. As we discussed earlier the loss and perplexity score shows the model's performance in the same manner. Decreasing the loss of a model is in favor of the model in the same way decreasing the perplexity also indicates the better performance of the model.

The following Fig4 shows the graph between perplexity score and loss. The following fig4 shows that there is a strong linear relationship between perplexity score and the loss value and the nature of the relationship is positive. 
\begin{figure}[h!]
	\caption{Training Results}
	\centering
	\includegraphics[width=0.61\textwidth]{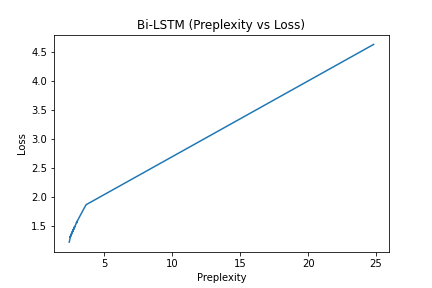}
\end{figure}
Almost, 90 percent of the total number of epochs there is a very straight and linear line. But during the remaining 10 percent of the epochs decrement in perplexity score becomes slower but not in loss value that why the slope during these epochs seems to be greater.

\subsection{Results over Test Dataset}
To evaluate the inclination of the model toward the ideology and style of a specific author requires a general type of intelligence, here we used perplexity score over a test set of the same author and different authors to evaluate.
\subsubsection{Inclination of the Model towards Ideology and Style}
In this section, we are going to evaluate the model from the perspective that how much the model is inclined towards the ideology and style of an author whose corpus is used in training. It is done by comparing perplexity scores obtained over the test dataset of the same author and different authors. As, the model is not able to manipulate the words through semantics, but can keep the track of probability of a character being next after a sequence of characters through perplexity score. So, the same ideology and style will try to reduce perplexity scores and vice versa.The following Fig5 shows diffferent experiements, each experiment is performed on a text chunk of same auhtor's test set and a text chunk from different author.

\begin{figure}[h!]
	\caption{Same author vs Different author(Test Set)}
	\centering
	\includegraphics[width=0.61\textwidth]{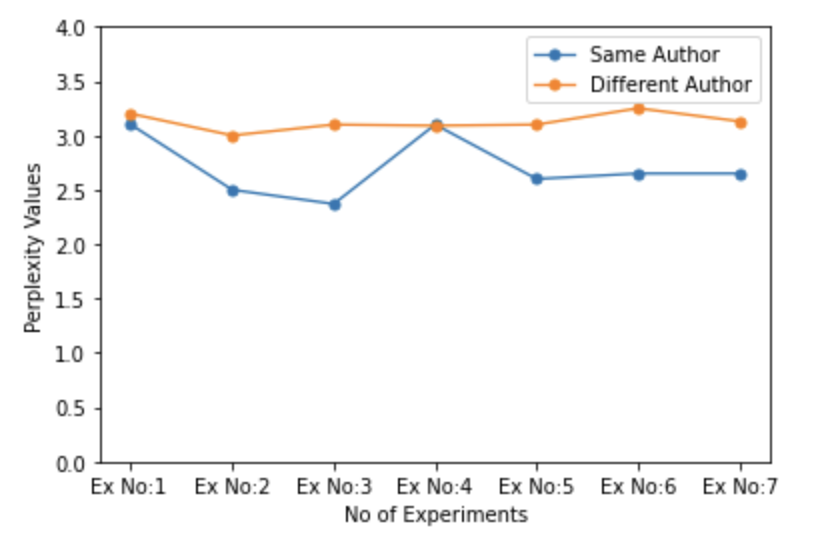}
\end{figure}

In fig5 it can be seen that most of experiment shows higher perplexity against different author becasue model is inclined towards ideology and style of a specific author when input text from corpus of different author is given to model, model tries to predict the characters according to idelogy and style of same author but perplexity is calculated according to different auhtor's corpus, due to contradiction in idelogy and style of both authors perplexity scores goes higheir.In case of same author, model predictions and acutal text have much similer ideology and style due to which perplexity scores are goes down.

\subsubsection{Loss and Perplexity over Testset}
In this section, we are going to visualize and explain the loss and perplexity scores over the test dataset with help of different experiments. The following Fig6 shows the comparison of loss and perplexity scores against each of the five experiments.

\begin{figure}[h!]
	\caption{Comparison of Loss and Perplexity}
	\centering
	\includegraphics[width=0.61\textwidth]{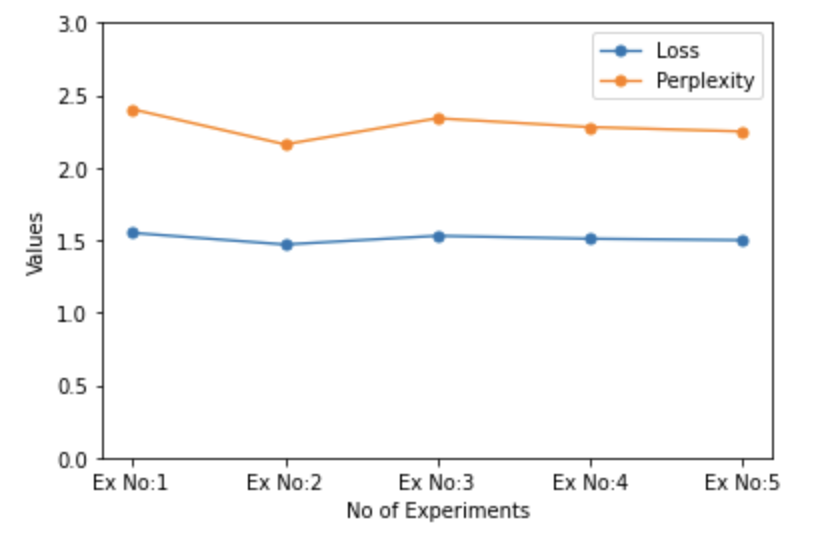}
\end{figure}

The above graph shows the loss and perplexity score over the test dataset against 5 different experiments. It can be seen that in each experiment both values seem to be very close to each other which shows that model is consistent in predicting the text in different experiments. In each experiment, different text chunks are taken from the test dataset, and it means that selecting the text chunk from the test dataset as input to the model does not matter for perplexity scores and loss value. In experiment no1 value of the loss is just above the 1.5 and the perplexity score is just above the 3. It means that having the loss of 1.5 defines the model as having 3 options averagely at each point of prediction. As the ideal case of perplexity, the score is having a value of 1, but it is very hard to achieve. That means the loss value should be reduced to around 0.2 to have a perplexity score of 1. Our model is based on character level predictions, it causes the non-dictionary words generation whenever the perplexity score is greater than 1. In case of having a perplexity score value of 1, it will finish the chances of predicting non-dictionary words. In experiment no3 loss value is the maximum which is 1.6 and the perplexity score for this experiment is 3.3, but it's not much greater than the others.

\subsubsection{Non-Dictionary Words as Predictions}
In this section, we going to discuss the percentage of non-dictionary words predicted by our model. In the following Fig7, the graph shows the percentage of non-dictionary words predicted by our model.

\begin{figure}[h!]
	\caption{Percentage of Non-Dictionary Words}
	\centering
	\includegraphics[width=0.61\textwidth]{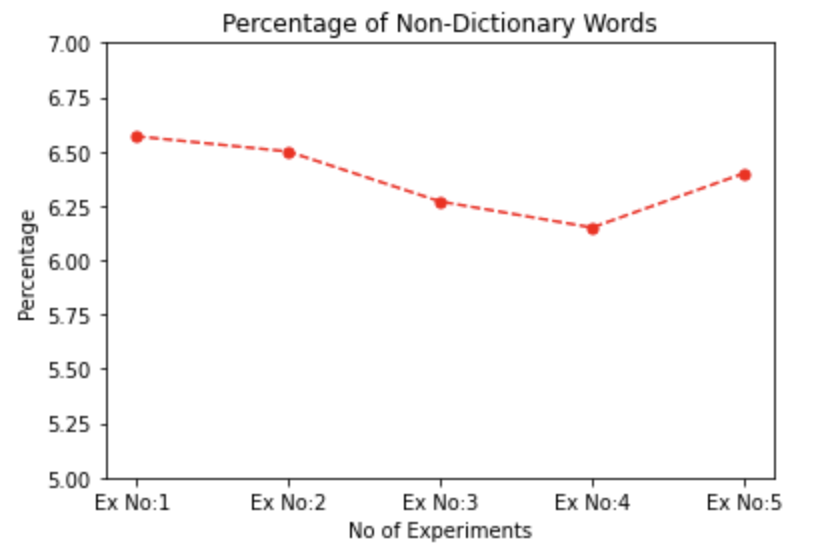}
\end{figure}

The above graph shows that in experiment no1 percentage of non-dictionary word prediction is just above the 7. In other experiments, this percentage is between 6 and 7. In the previous section, we briefly explain the cause of non-dictionary word prediction and explain how much work is required to reduce this percentage to zero.

\subsubsection{Analysis and Discussion}
In this section, we will analyze and discuss the results of our model on the test dataset. The text generated by the model against the sequence of words given to the model as the input seems to be based on the ideology and style of the author. To analyze the generated output requires a comprehensive study of the training corpus of the author.However, the inclination of the model towards ideology and the style of the specific author can be analyzed by comparing the perplexity scores on a test set of the same author and different authors. In section 4.3.1,graph in fig5 shows that the perplexity score on a test set of the same author is less than a different author in most of the experiments,it happened because when the model is tested on a different author's test set there is a difference between style and ideology, model is trying to predict and actual style and ideology(model should predict). That is the reason the perplexity score goes up.On the other hand, perplexity scores for test set of same auhtor are less becasue style and ideology, model is trying to predict and model should predict is same. That's why perplexity scores goes down in this case.So, our trained model is good at capturing style and ideology of specific auhtor.

\subsection{Comparison with other Models}
In this section, we are going to explain the results generated by other models on our test dataset by comparing each other. The models we used for comparison are based on RNN, LSTM(unidirectional), RNN+LSTM, and Transformers. The following Table 3 shows the perplexity scores generated by the mentioned models over the test dataset.
\begin{table}[htb]
	\centering
	\begin{tabular}{lc}
		\hline
		\textbf{Model} &  \textbf{Perplexity}\\
		\hline
		RNN   & 3.80  \\
		\hline
		LSTM(Uni)  & 3.20  \\
		\hline
		RNN+LSTM   & 2.73  \\
		\hline
		Transformer   & 1.67  \\
		\hline
	\end{tabular}
	\caption{Comparison with other Models}
	\label{tab:accents}
\end{table}

The above table shows the performance of each model as  perplexity scores.In the table, the loss is not mentioned but is explained in this paragraph. Now, we will discuss them one by one. First of all, there is a Recurrent Neural Network(RNN) based neural language model. It is trained on the same dataset and tested against the same test dataset. This model achieved a minimum of 1.95 as a loss value against which the perplexity score is 3.80. The second model is Long Short Term Memory(LSTM-Uni), which achieves a minimum of 1.79 loss value and perplexity score of 3.20. The third neural language model is based on the RNN and LSTM. It gives a minimum loss value of 1.65 and 2.73 as perplexity values. The last one is the Transformer based language model which achieves a loss value of 1.29 and a perplexity score of 1.67. In the comparison table, it can be seen that Transformer outperforms, but if we compare this model with our model although the perplexity score achieved by this model is better in the extraction of ideology and style this model does not seems to be good because of word-level predictions. So, we can say that Bidirectional LSTM outperforms overall in the context of ideology and style.
\section{Limitations}
This research work does not focus on generating only meaningful words, due to character-level predictions sometimes
the model generates non-dictionary words.It happened due to perplexity score is higher than 1. It is hard to reduce the perplexity scores of a model to 1, but it grantee the generation of only dictionary words.  The second limitation of this research is a model only can digest a limited amount of ground facts dataset in a reasonable amount of time during
training.The main reason of model to digest a limited amount of ground facts dataset is contradiction check between ground facts and author's corpus.Because before sending each text chunk of ground facts datset to the model for training, it sent to pre-trained model for contradiciton check with chunks of author's datset.It is done for every chunk of ground facts datset that's why model is limited to digest small volume of ground facts datset.
\section{Conclusion}
Our thinking about any individual is based on their ideology, even we can predict the response of anyone we used to talk. This is because of training of our mind on the previous response of the same individual against different actions. In the same way, ideology and responses can be extracted from text using a neural language model, after extraction this ideology and style new text can be generated against a sequence of input words. Our model ables us to extract the ideology and style of a specific author by training the model on a corpus of the same author. To generate text on ground truth, a model needs to train on ground truth corpus as well but only those chunks are used which have no contradiction with the author's corpus because the goal is to generate a model inclined towards the ideology and style of the author. To stop generating non-dictionary words loss should be reduced to around 0.23 to have a perplexity score of 1. This will make the model ables to predict only dictionary words.
\section{Future Work}
This works gives two different ways for future research. The first one is how the model can be trained on the maximum ground truth corpus without the inclination of the model towards it. It can be done with a logical contradiction check with feasible requirements of computational power. The second one is reducing the model's loss to 0.23 to make perplexity 1. As it will make a model generate only dictionary words.
\section*{Acknowledgement.}
This work was supported by AIM Lab, National University of Computer and Emerging Sciences.
%Bibliography
\bibliographystyle{unsrt}  
\bibliography{Cloning_Ideology_and_Style}

\end{document}